\documentclass[sigconf,nonacm,natbib=true]{style/acmart}

\usepackage[]{svg}
\usepackage{multirow}
\usepackage{caption}
\usepackage{subcaption}
\usepackage{algorithm}
\usepackage{algpseudocode}
\usepackage{afterpage}
\usepackage{tablefootnote}
\usepackage{appendix}
\usepackage{float}

\usepackage{graphicx}
\usepackage{etoolbox}
\usepackage{acronym}

\usepackage{xcolor}

\definecolor{codegreen}{rgb}{0,0.6,0}
\definecolor{codegray}{rgb}{0.5,0.5,0.5}
\definecolor{codepurple}{rgb}{0.58,0,0.82}
\definecolor{backcolour}{rgb}{0.95,0.95,0.92}
\usepackage{listings}
\lstdefinestyle{mystyle}{
    backgroundcolor=\color{backcolour},   
    commentstyle=\color{codegreen},
    keywordstyle=\color{magenta},
    numberstyle=\tiny\color{codegray},
    stringstyle=\color{codepurple},
    basicstyle=\ttfamily\footnotesize,
    breakatwhitespace=false,         
    breaklines=true,                 
    captionpos=b,                    
    keepspaces=true,                 
    numbers=left,                    
    numbersep=5pt,                  
    showspaces=false,                
    showstringspaces=false,
    showtabs=false,                  
    tabsize=2
}
\newbool{showcomments}

        \boolfalse{showcomments}

    \newcommand{\pinaforecomment}[4]{%
    \ifbool{showcomments}{%
        \colorbox{#1}{\textcolor{#4}{\parbox{.8\linewidth}{#2: #3}}}%
    }{}%
}

\begin{document}


\graphicspath{ {figures/}{auto_commit_fig/}{auto_fig/} }


\newcommand{\osullikomment}[1]{\pinaforecomment{green}{Kent}{#1}{black}}
\newcommand{\nrscomment}[1]{\pinaforecomment{violet}{Nicole}{#1}{white}}
\newcommand{\nrcomment}[1]{\pinaforecomment{blue}{Nandini}{#1}{white}}

\NewDocumentCommand{\rot}{O{45} O{4.25em} m}{\makebox[#2][c]{\rotatebox{#1}{#3}}}%


\title{DistRAG: Towards Distance-Based Spatial Reasoning in LLMs}

\author{Nicole R. Schneider}
\email{nsch@umd.edu}
\affiliation{%
  \institution{University of Maryland}
  \city{College Park}
  \country{USA}
}

\author{Nandini Ramachandran}
\email{nandinir@terpmail.umd.edu }
\affiliation{%
  \institution{University of Maryland}
  \city{College Park}
  \country{USA}
}

\author{Kent O'Sullivan}
\email{kosu0918@uni.sydney.edu.au}
\affiliation{%
  \institution{University of Sydney}
  \city{Sydney}
  \country{Australia}
}

\author{Hanan Samet}
\email{hjs@cs.umd.edu}
\affiliation{%
  \institution{University of Maryland}
  \city{College Park}
  \country{USA}
}

\begin{abstract}

Many real world tasks where Large Language Models (LLMs) can be used require spatial reasoning, like Point of Interest (POI) recommendation and itinerary planning.
However, on their own LLMs lack reliable spatial reasoning capabilities, especially about distances.
To address this problem, we develop a novel approach, DistRAG, that enables an LLM to retrieve relevant spatial information not explicitly learned during training.
Our method encodes the geodesic distances between cities and towns in a graph and retrieves a context subgraph relevant to the question.
Using this technique, our method enables an LLM to answer distance-based reasoning questions that it otherwise cannot answer.
Given the vast array of possible places an LLM could be asked about, DistRAG offers a flexible first step towards providing a rudimentary `world model' to complement the linguistic knowledge held in LLMs.
\end{abstract}

\maketitle

\section{Introduction}
\label{section:introduction}

Large Language Models (LLMs) can be applied to various tasks, including ones grounded in the real world that require spatial understanding to infer information not learned during training.
This requirement is particularly notable for tasks like Point of Interest (POI) recommendation and itinerary or travel planning, where correct spatial reasoning is needed to formulate a reasonable response and avoid sending users on a circuitous route~\cite{Schneider2025}.
Although previous work has indicated LLMs inherit some geospatial knowledge through training~\cite{Bhandari2023}, further work has shown they cannot correctly answer questions that require inferring distances between places~\cite{Osullivan2024}, which is critical for spatial tasks.

The challenge of applying LLMs to tasks involving geospatial reasoning is further complicated by the lack of representation of lesser-known places in training data.
Many obscure places have long, complex names that are out-of-vocabulary for LLMs, making it even harder for LLMs to associate them with their physical location in the world and reason about them.
Recent work has seen success using Retrieval Augmented Generation (RAG) to provide an LLM with domain-specific information that did not appear in its training corpus~\cite{Lewis2020}.
Given the vast array of possible places an LLM could be asked to reason about, RAG offers a flexible first step towards providing an LLM with a rudimentary `world model' to complement its vast linguistic knowledge.

We develop a novel approach, \emph{DistRAG}~\footnote{\url{https://github.com/nani-r/dist-rag}}, that allows an LLM to retrieve spatial information that it has not memorized during training.
We encode the geodesic distances between cities and other locations in an attributed graph, and use a retriever to select a context subgraph to provide the LLM based on the question in the prompt.
Using this technique, our method successfully answers a range of distance-based spatial reasoning questions that the same base LLM without DistRAG cannot answer.
DistRAG represents a step towards grounding LLMs in the real world. 
It provides them with a means to access a simplified world model that can be updated without retraining the model weights.
Critically, as language models play an increasing role in information access, DistRAG provides an avenue for LLMs to answer questions about places that are not popular enough to appear in training data, which paves the way for fairer, more geographically accurate responses.
\section{DistRAG}
\label{section:background}
\normalsize

We design DistRAG using ideas from GraphRAG~\cite{Peng2024,Wu2023,Sen2023} and spatial reasoning~\cite{Schneider2024, Osullivan2024}.

\subsection{Background}
RAG techniques face challenges with hallucination, where the model generates spurious tokens not grounded in the underlying knowledge store, and faulty retrieval, where the model generates an inaccurate response faithfully grounded in incorrectly retrieved content.
For document-based RAG approaches, careful selection of document snippets is critical for an accurate response~\cite{Procko2024}.
The same is true of geospatial data, which is naturally stored as a graph, with edge attributes explicitly capturing relationships between places.
Some RAG approaches allow retrieving over graphs, typically Knowledge Graphs (KGs)~\cite{Wu2023,Sen2023}.
Successful retrieval requires determining the appropriate subgraph to provide as context, without including unnecessary nodes and edges, which is an NP-hard problem~\cite{Procko2024}.
Since spatial graphs are dense, choosing the appropriate subgraph without including extraneous information is a challenge.
We explore this through two different retrieval methods.

\subsection{Method}
DistRAG has two critical components, the spatial knowledge store and the retriever, which together provide the LLM with the appropriate spatial context for the question.
Below we describe these components, which we construct using \textit{LangChain}.~\footnote{\url{https://python.langchain.com/docs/introduction/}} 

\textbf{Spatial Graph Construction}
The spatial knowledge store is maintained as a graph consisting of triples where Nodes are string representations of city names, and an Edge contains the distance between the two Nodes it connects.
For example, the triple
\begin{verbatim}
    ("Newcastle, NSW", "Sydney, NSW", "160 km")
\end{verbatim}
represents the distance relationship between Newcastle, Australia and Sydney, Australia.
The graph is populated with edge attributes containing the distances between the nodes they connect.
Cities in the data store are selected from Open Streetmap (OSM)~\cite{Haklay2008} using the OSMnx library~\cite{Boeing2024}.

\textbf{Retriever}
The retrieval component of our method selects the geographically relevant subgraph using one of two methods:
\begin{enumerate}
    \item Vector similarity between graph triples and the prompt, or
    \item SPARQL query generation based on the prompt.
\end{enumerate} 

For the vector similarity approach, we vectorize the prompt and triples comprising the graph and use Facebook Artificial Intelligence Similarity Search (FAISS)~\cite{Douze2024} to retrieve a list of triples that are similar to the prompt.
This technique allows for flexible matching for instances of polyonymy, where a single location is known by many names, which may be captured in the vector embedding space. 
We extract the top $k$ triples from the graph and construct a natural language representation of the relevant subgraph, which is provided to the LLM using Prompt Template 1.
We set the hyperparameter value $k$ = 10 for evaluation, since this was the best-performing setting, balancing noisiness and sparsity in the retrieval process.
\begin{lstlisting}[title=Prompt Template 1 - Vector Similarity Retrieval]
 You are an assistant with access to a graph database 
 containing distances between cities. Your task is to 
 extract and return the distance between the cities 
 mentioned in the question. Given the following question, 
 provide a direct answer. Only return the distance (km) 
 or the city name (text) as requested in the question. Do 
 not include any additional information including units.
 Question:  {question}
 Context:  {graph_context}
\end{lstlisting}

For the SPARQL query approach, we instruct the retriever to write a SPARQL query appropriately capturing the question.
The query is executed against the spatial graph, which is stored in Resource Description Framework (RDF) format.
We use Prompt Template 2 to instruct the retriever.

\begin{lstlisting}[title=Prompt Template 2 - SPARQL Retrieval]
 Convert the following natural language question into a 
 SPARQL query for an RDF graph. The RDF graph follows 
 this structure:
- Cities are identified by URIs in the namespace <http://example.org/cities#>.
- The predicate `ns1:distanceTo` represents relationships between cities.
- The predicate `ns1:distance` represents the numeric distance between cities.
 Example RDF:
  ns1:Adelaide a ns1:City ;
  ns1:distanceTo 
  [ ns1:destination ns1:Perth ; ns1:distance 2135 ],
  [ ns1:destination ns1:Launceston ; ns1:distance 1039 ],
  [ ns1:destination ns1:Cairns ; ns1:distance 2119 ],
  [ ns1:destination ns1:Ipswich ; ns1:distance 1571 ],
  [ ns1:destination ns1:Mount_Isa ;
 Generate a SPARQL query to answer the question. The 
 result should either be a distance or a city name 
 depending on the question. Note: If there are spaces in 
 city names, you will have to replace them with 
 underscores (_)
 ### Expected SPARQL Query: 
       PREFIX ns1: <http://example.org/cities#>
 ### Question: {question}
 SPARQL Query:
\end{lstlisting}

\section{Experiments}
\label{section:experiments}

We evaluate the DistRAG method against a baseline LLM on distance-based reasoning questions of varying difficulty.

\subsection{Datasets}
We perform an empirical evaluation using a dataset of 60 distance-based spatial reasoning questions, varying in difficulty from Easy to Medium, to Difficult. 
We construct all of the questions using towns and cities in Australia, chosen because it is an English-speaking country with Anglicized place names likely to be in the vocabulary of an LLM and Indigenous place names that are less likely to be in the vocabulary. 
Further, Australia is a large country with dispersed, dense settlements that allow for testing distance-based geospatial reasoning across varying scales.

\textbf{Easy.} The Easy questions take the following form, where \textbf{A} and \textbf{B} are place names:
    \textit{``What is the distance between A and B?''}
These questions are easy because they can be answered using the value on edge between A and B, which are both named in the question.

\textbf{Medium.} The Medium questions take the following form, where \textbf{A} is a place name:
    \textit{``What is the distance between A and its closest city?''}
These questions are of medium difficulty because they require one `hop' of reasoning, involving evaluating every node's relationship to A and selecting the one whose distance from A is the smallest. 

\textbf{Difficult.} The Difficult questions take the following form, where place names \textbf{A} and \textbf{B} are provided, and the model produces a place name with an appropriate distance to \textbf{C}:
     \textit{``The distance from A to B is similar to the distance from C to what other city or town?''}
These questions are difficult because they require two `hops' of reasoning; finding the distance, say $x$, between A and B, and evaluating neighbors of C to determine whose distance from C is closest to $x$.

\subsection{Baseline Model}
We compare both versions of DistRAG (vector similarity-based and SPARQL query-based retrieval) to a baseline LLM (GPT-4-0613\footnote{\href{https://platform.openai.com/docs/models\#gpt-4-turbo-and-gpt-4}{https://platform.openai.com/docs/models\#gpt-4-turbo-and-gpt-4}}) to determine the effect of our RAG methods versus the base LLM alone.
We use the following prompt template for the baseline LLM.
\begin{lstlisting}[title=Prompt Template 3 - Baseline LLM]
 Given the following question, provide a direct answer. 
 Only return the distance (km) or the city name (text) as 
 requested in the question. Do not include any additional 
 information including units. Do not provide any 
 additional context except the answer.
  Question:  {question}        Answer:
\end{lstlisting}

\subsection{Evaluation Metrics}
For Easy and Medium questions we calculate error as the difference in Kilometers between the model response and the actual geodesic distance between the relevant cities.
For Difficult questions, we geocode the city returned by the model and calculate error as the difference between the target distance and the geodesic distance between the model-produced city and the reference city.
For all questions we compute Mean Squared Error (MSE):
\(MSE = \frac{1}{n} \sum_{i=1}^{n} (y_i - \hat{y}_i)^2,\) 
where 
$n$ is the number of data points,
$y_i$ is the true distance for the $i$-th data point,
and
$\hat{y}_i$ is the predicted value based on model output for the $i$-th data point.
We record response time from query issuance to response~\footnote{For DistRAG includes retrieval time but not time building knowledge store, a one-time upfront cost}.

\subsection{Ablation Study}
We further study the effect of the 
knowledge store completeness on DistRAG performance.
We evaluate robustness by measuring performance as the density of the spatial knowledge store varies from sparse to fully connected.
This test captures how DistRAG performs under sparse retrieval, simulating when there is missing or incomplete spatial information available.

\section{Results}
\label{section:results}

\textbf{Performance Results.}
We report the MSE for DistRAG and the baseline method on the Easy, Medium, and Difficult datasets in Table~\ref{tab:results}.
We find that DistRAG outperforms the baseline on Easy and Medium spatial reasoning questions with a 100\% decrease in MSE for both variants on Easy questions and a 90\% and 100\% decrease in MSE on Medium questions for the vector similarity and SPARQL variants, respectively. 
On the Difficult dataset, all three methods perform poorly, with DistRAG-SPARQL abstaining from answering any questions, and gpt-4 alone performing the best, but with high error rates.
After further providing DistRAG-SPARQL with a template of the query structure, it achieves 0 MSE, correctly answering every Difficult question, indicating that the model is capable of constructing the queries needed to answer the questions, but that it cannot perform the complex reasoning to arrive at the correct query structure on its own.

\begin{figure}
    \centering
    \includegraphics[width=0.95\columnwidth]{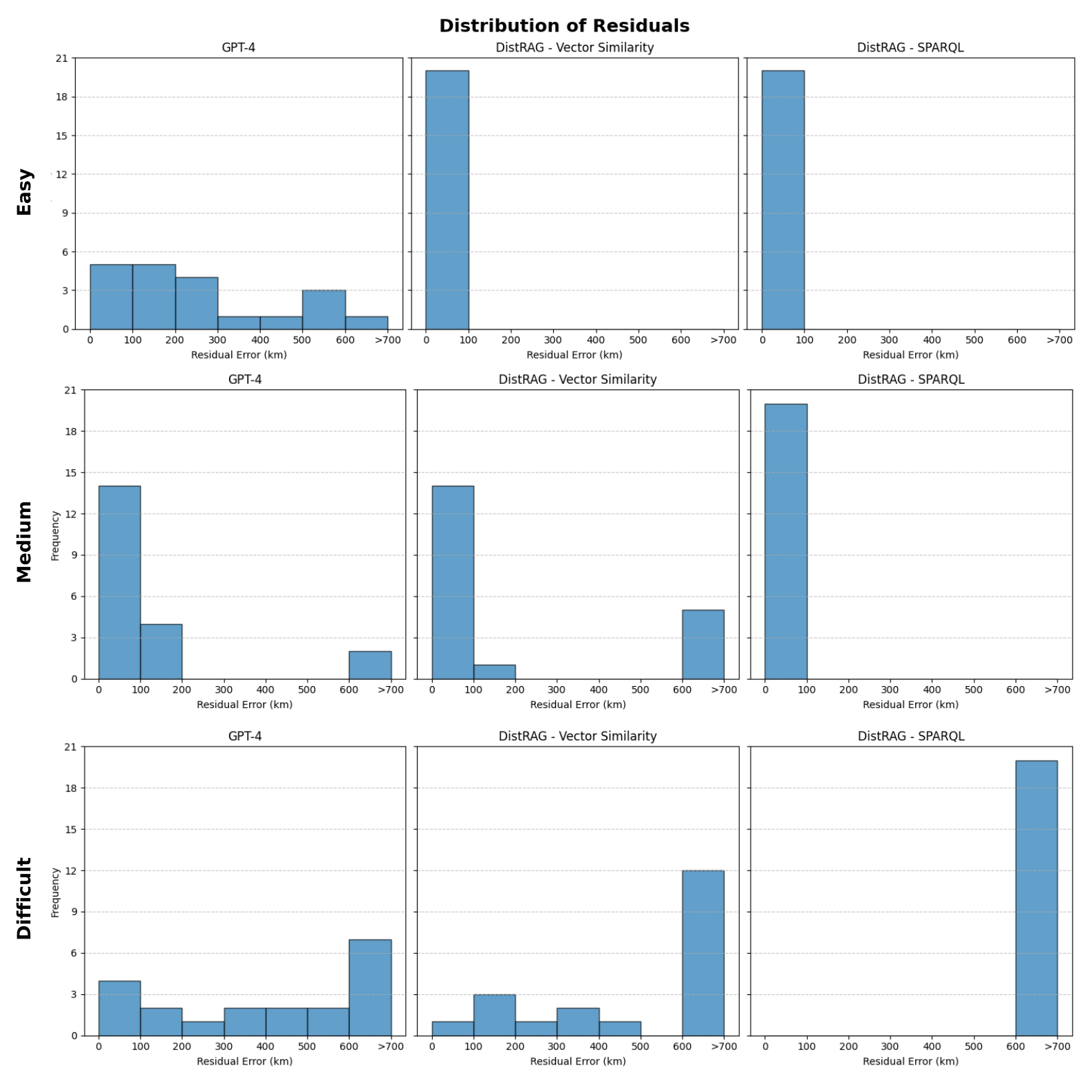}
    \caption{Residual errors of DistRAG and baseline method by dataset. Abstention is binned with errors > 700km.}
    \label{fig:combined-results-plot}
\end{figure}

We further visualize the residual errors in Figure \ref{fig:combined-results-plot}.
For the baseline method, errors tend to be spread evenly, ranging from less than 100km to greater than 700km, indicating the model is frequently guessing answers without regard to their spatial relationship to the query city.
The two DistRAG variants tend to have smaller residuals, mostly less than 100km, except on the Difficult questions, where they have large residuals or fail to answer many of the questions.
These findings suggest that more work is needed to address complex reasoning tasks, since the Difficult questions required a level of reasoning that no method was able to achieve consistently.
We also observe that SPARQL-based retrieval has an ``all or nothing'' success rate based on whether the correct query is written.
In fact, it abstained from answering any of the Difficult questions, meaning that the answers it did give (for Easy and Medium questions) were correct every time.
In terms of response time, the baseline method took a total of 61.9 seconds to answer all 60 questions, whereas the DistRAG method took 137.1 seconds and 274.1 seconds for the vector similarity and SPARQL variants, respectively.

\begin{figure}
    \centering
    \includegraphics[width=\columnwidth]{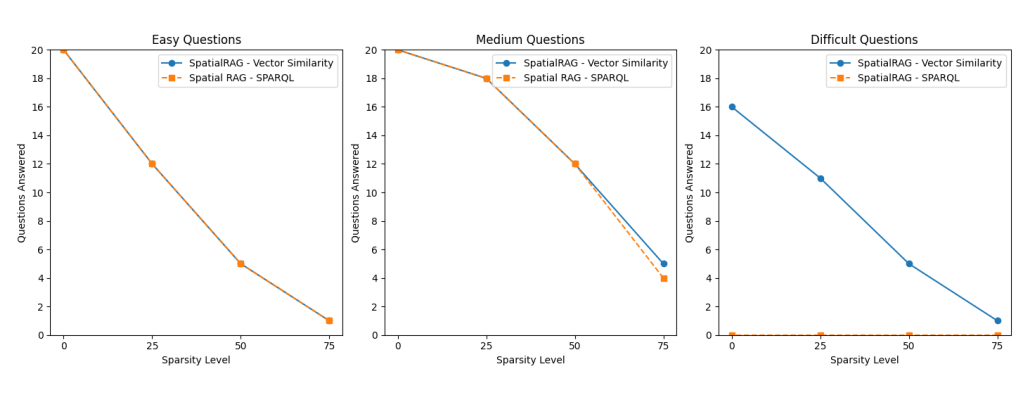}
    \caption{Ablation Results. Response rate by question type for DistRAG with varying  knowledge store sparsity. A sparsity of 0\% is fully connected. Larger sparsity represents a greater percentage of missing edges (less complete knowledge store).}
    \label{fig:sparsity-plots}
\end{figure}

\textbf{Ablation Results.}
Figure \ref{fig:sparsity-plots} shows the rate of abstention of \mbox{DistRAG} across varied spatial data stores, ranging from fully connected graphs to 75\% sparse graphs (meaning only 25\% of the edges are retained).
Unsurprisingly, the rate of refusal to answer increases as the knowledge store becomes sparser (less complete), with a reasonable level of tolerance to a small percentage of information missing (i.e. 25\% of edges missing).
Further, the SPARQL-based retrieval method shows more brittleness to sparsity, refusing to answer more frequently than the vector similarity-based alternative, although we observe that it is correct every time it does answer.

\begingroup
\small
\begin{table}[ht]
    \centering
    \begin{tabular}{l|lrr}
        \textbf{Dataset} & \textbf{Method}            & \textbf{MSE} $\downarrow$    & \textbf{Abstains} $\downarrow$    \\
        \hline
        Easy         & gpt-4                      & $1.01\times 10^{5}$   & -     \\
        ~                & \textbf{DistRAG-Vector} & \boldmath{$0$} & - \\
        ~                & \textbf{DistRAG-SPARQL} & \boldmath{$0$} & - \\
        
        Medium         & gpt-4                      & $3.07\times 10^{4}$ & 1/20       \\
        ~                & DistRAG-Vector & $3.04\times 10^{3}$ & - \\
        ~                & \textbf{DistRAG-SPARQL} & \boldmath{$0$} & - \\
        Difficult       & \textbf{gpt-4}        & \boldmath{$4.11\times 10^{7}$} & 2/20          \\
        ~                & DistRAG-Vector & $8.34\times 10^{7}$ & 1/20 \\
        ~                & DistRAG-SPARQL & $-$ & 20/20\\
    \end{tabular}
    \caption{Error results for DistRAG with vector similarity and SPARQL-based retrieval vs. baseline method gpt-4. Abstains indicates how many questions the model did not answer.}
    \label{tab:results}
\end{table}
\endgroup

\section{Related Work}
\label{section:related}

Work like GeoLLM~\cite{Manvi2024} and Spatial-RAG~\cite{Yu2025} design spatially-aware LLMs that can answer geospatial prediction questions, like determining population variables (GeoLLM), or generating POI recommendations (Spatial-RAG) but they do not address reasoning about the relative distances between places.
NALSpatial, uses entity mapping rules to construct executable structured spatial queries, like range queries, with 95\% translatability, but this also does not directly apply to general distance-based reasoning questions~\cite{Liu2023}.
Other works envision the development of geo-foundation models without making concrete progress towards enabling geospatial reasoning abilities~\cite{Bhandari2023, Qi2023}.

Many RAG approaches incorporate graphs or graph databases~\cite{Procko2024,Peng2024}, with applications ranging from specific domains of interest to broad open-domain question answering. 
Some works encode the graph or knowledge store using natural language~\cite{Fatemi2023, Ye2023}, finding that the graph encoding, the type of graph task, and even the structure of the graph influence LLM performance~\cite{Fatemi2023}.
A similar approach has been successfully used on knowledge graphs for open-domain question answering by re-ranking retrieved graph triples to avoid discarding relevant information not part of the local subgraph~\cite{Li2023}.
However, it is limited to multiple choice questions~\cite{Li2023} and does not account for the unique characteristics of spatial data, which is naturally dense, since distance relationships can be determined between any given pair of locations.
Finally, several works store the graph in RDF and convert natural language questions into SPARQL queries, finding the task challenging, especially for compound queries~\cite{Dubey2016,Shaik2016,Yin2021}.
Our results support this idea, since our DistRAG-SPARQL variant abstained from answering difficult questions that required complex queries.

\section{Conclusion}
\label{section:conclusion}
\normalsize

We present DistRAG, a method for improving the spatial reasoning ability of LLMs using graph retrieval over a spatial knowledge store.
We demonstrate DistRAG's performance over an LLM on distance-based spatial reasoning questions and show that the method is robust to missing information in the knowledge store.
This work presents a significant step towards enabling LLMs to support applications that require reasoning about distance. 
It can be expanded to include other spatial relationships
and improve reasoning on complex questions, which prove difficult even with RAG.

\begin{acks}
This work was sponsored in part by the \grantsponsor{1}{NSF}{} under Grants \grantnum{1}{IIS-18-16889}, \grantnum{1}{IIS-20-41415}, and \grantnum{1}{IIS-21-14451}.
\end{acks}

\bibliographystyle{style/ACM-Reference-Format.bst} 
\bibliography{main.bib} \label{bibliography}

\end{document}